%% file: main.tex
\documentclass[10pt,twocolumn,letterpaper]{article}

\usepackage[pagenumbers]{cvpr} 

\usepackage{graphicx}
\usepackage{amsmath,bm}
\usepackage{amssymb}
\usepackage{booktabs}
\usepackage{import}
\usepackage{comment}
\usepackage{nicefrac}
\usepackage{array}
\usepackage{soul}
\usepackage[pagebackref,breaklinks,colorlinks]{hyperref}

\usepackage[capitalize]{cleveref}
\crefname{section}{Sec.}{Secs.}
\Crefname{section}{Section}{Sections}
\Crefname{table}{Table}{Tables}
\crefname{table}{Tab.}{Tabs.}

\def\cvprPaperID{****} 
\def\confName{CVPR}
\def\confYear{2024}

\newcommand{\name}{\textsf{DiffSBR}\xspace}
\begin{document}

\title{Differentiable Radio Frequency Ray Tracing for Millimeter-Wave Sensing}

\author{Xingyu Chen$^{1,2}$, Xinyu Zhang$^1$, Qiyue Xia$^3$, Xinmin Fang$^2$, Chris Xiaoxuan Lu$^3$, Zhengxiong Li$^2$\\\\
$^1$UC San Diego \;  $^2$University of Colorado Denver \; $^3$University of Edinburgh \\}


\maketitle

\begin{abstract}

Millimeter wave (mmWave) sensing is an emerging technology with applications in 3D object characterization and environment mapping. However, realizing precise 3D reconstruction from sparse mmWave signals remains challenging. Existing methods rely on data-driven learning, constrained by dataset availability and difficulty in generalization. We propose DiffSBR, a differentiable framework for mmWave-based 3D reconstruction. DiffSBR incorporates a differentiable ray tracing engine to simulate radar point clouds from virtual 3D models. A gradient-based optimizer refines the model parameters to minimize the discrepancy between simulated and real point clouds. Experiments using various radar hardware validate DiffSBR's capability for fine-grained 3D reconstruction, even for novel objects unseen by the radar previously. By integrating physics-based simulation with gradient optimization, DiffSBR transcends the limitations of data-driven approaches and pioneers a new paradigm for mmWave sensing.

\end{abstract}


\subimport{Chapter/}{introduction}
\subimport{Chapter/}{related_work}

\subimport{Chapter/}{system_design}

\subimport{Chapter/}{implementation}

\subimport{Chapter/}{experiment}

\subimport{Chapter/}{discussion}
\subimport{Chapter/}{conclusion}

{\small
\bibliographystyle{ieee_fullname}
\bibliography{egbib}
}

\end{document}

%% file: Chapter/introduction.tex
\section{Introduction}
\label{sec:introduction}

Millimeter wave (mmWave) sensing is a burgeoning field with vast implications for surveillance, security \cite{appleby2007millimeter,li2022spiralspy}, autonomous navigation \cite{nguyen2020dronescale, nolan2021ros}, etc. MmWave radar sensors in particular have gained immense popularity in recent years, owing to their capability to discern objects' range and angles and even generate point clouds \cite{qian20203d}. Robustness against lighting and atmospheric conditions primes mmWave radar for roles where conventional cameras and lidar falter \cite{shen2022sok, chen2023metawave}. Despite such advantages, realizing precise \textit{3D object characterization} with mmWave technology—a process critical for understanding complex scenes and behaviors—has been constrained by the limited spatial resolution of available mmWave sensors.

Recently proposed data-driven mmWave sensing models \cite{zhao2019through,guan2020through,xue2021mmmesh,ren20213d,jiang2020towards}, while offering potential for object classification, encounters barriers when advancing toward the nuanced goal of \textit{3D mesh reconstruction}. These barriers include the heavy reliance on large, diverse datasets, the difficulty in generalizing beyond learned object types, and inability to adapt to new radar hardware without extensive retraining.

In this work, we challenge the status quo by introducing \name, a new approach anchored by a differentiable radio frequency (RF) ray-tracing simulator that enables gradient-based 3D reconstruction. Central to our contribution is the advancement of a differentiable RF simulation capable of bridging the gap between sparse mmWave radar point clouds and detailed 3D object geometries. This novel simulator allows for the backpropagation of loss scalar, facilitating the fine-tuning of simulated parameters to mimic the real-world radar observations.

By harnessing the power of differentiable programming within the RF domain, \name sets a precedent in mmWave-based 3D object characterization. \name transcends the constraints of data-hungry methods by allowing for the characterization of objects previously unseen by the radar, thus minimizing the need for exhaustive data collection. Our experiments on a variety of radar platforms and real-world scenes reveal that \name not only achieves remarkable accuracy in reconstructing object shapes and sizes, but also demonstrates an impressive ability to infer the 3D mesh of novel objects directly from sparse mmWave signals.

The key contributions of \name are two folds. First, we introduce an RF ray tracing simulator that can represent the temporal-frequency patterns of mmWave radar signals, along with their spatial propagation and interaction with objects. We design new mechanisms to make the entire simulator differentiable, so that it can be incorporated into a wide range of RF optimization problems. 
Second, leveraging the differentiable RF simulator, we formulate the radar-based 3D reconstruction as a gradient-driven optimization framework that matches virtual objects to measured radar point clouds. 
This framework departs from recently proposed data-driven approaches as it is easy to generalize and requires no radar training data. Comprehensive experiments on real radar hardware and in diverse environments demonstrate the effectiveness of our methods. 

%% file: Chapter/related_work.tex
\section{Related Work}

\subsection{Millimeter-Wave (mmWave) Sensing}

MmWave sensing technologies recently garnered substantial interest in the domain of machine perception \cite{qian20203d,qian2021robust}, largely attributed to their resilience under challenging environmental conditions, e.g., low light, smoke, rain, snow and fog \cite{chen2023metawave, shen2022sok, wei2022mmwave}.   
Commercial mmWave automotive radar sensors can easily achieve multi-cm range (depth) resolution, owing to their high time resolution.   
However, their angular resolution is constrained by the antenna aperture \cite{mmWaveSenseBook}, which is directly proportional to the number of antenna elements -- analogous to the pixel count in a camera. 
Consequently, while these sensors can generate 3D point clouds, the resulting data points are notably sparse, typically amounting to mere dozens of points \cite{qian20203d}. 

Earlier studies of mmWave-based automotive perception primarily explored mmWave radars for obstacle detection \cite{sugimoto2004obstacle, wei2022mmwave}. More recent applications of mmWave sensing are imitating visual perception capabilities, such as gesture and posture tracking \cite{lien2016soli, kong2022m3track, li2022towards, lien2016soli}. Despite these advancements, existing mmWave-based object characterization models predominantly rely on data-driven black-box inference or black-box optimization \cite{zhao2019through,guan2020through,xue2021mmmesh,ren20213d,jiang2020towards}, which suffers from generalization due to (i) highly diverse radar hardware and (ii) lack of large, diverse radar datasets. An RF simulator tailored for mmWave signals can mitigate such limitations. More importantly, the simulator must be differentiable so as to seamless integrate with existing neural network models or gradient-based optimization frameworks. \name marks an important step in filling this gap. 

\subsection{Computational Electromagnetics}
Computational electromagnetics (CEM) has emerged as a powerful tool for simulating RF propagation and scattering. CEM techniques numerically solve Maxwell's equations to model electromagnetic wave interactions with objects and environments.
Historically, CEM relied on frequency and time domain methods, such as the finite-difference time-domain (FDTD) \cite{mohammadian1991computation} technique. The finite element method (FEM) \cite{hiptmair2002finite} and the method of moments (MoM) \cite{harrington1996field} have also seen extensive applications, particularly in antenna design. More recently, learning-based techniques, including neural networks \cite{mishra2002overview} and Gaussian processes \cite{wang2021deep}, have been explored for surrogate modeling. Conventional CEM methods often face restrictions in handling large simulation domains due to computational intensities. Contemporary research has leaned towards ray tracing for efficient large-scale propagation modeling \cite{he2018design, chen2023rfgenesis}.

To support cutting-edge wireless applications, like ambient computing or metamaterial design and intricate sensing \cite{chen2023metawave, yang2016programmable, zheng2017ultra}, optimization-based methods are essential. Yet, many of the existing techniques fall short, either due to their non-differentiable nature or because their computational overheads render them unsuitable for iterative processes. In this context, \name emerges as a flexible computational electromagnetics simulator, tailor-made for optimization-centric tasks. This facilitates a more seamless fusion of electromagnetic waves with deep learning models and enables the execution of gradient-driven optimization.

\subsection{Neural and Differentiable Rendering}

Neural rendering has seen rapid progress in recent years. Early works focused on neural techniques for novel view synthesis from a set of input views. These methods train networks to implicitly represent 3D scenes and render novel views through volumetric ray marching \cite{mildenhall2020nerf}. While able to generate high-quality results, they lack an explicit 3D representation and differentiability.
More recent works have focused on building differentiable renderers to enable end-to-end training for 3D reconstruction and novel view synthesis \cite{kato2020differentiable, petersen2022gendr, loubet2019reparameterizing}. These differentiable renderers approximate the traditional graphics pipeline, enabling gradient-based optimization of 3D representations like meshes, point clouds, or implicit functions. However, current differentiable rendering techniques predominantly focus on the visual scenes. \name aims to extend the principles of differentiable rendering into the RF domain. This expansion promises potential benefits for a multitude of applications, such as radar-based human activity recognition, autonomous driving, and programmable environment based on metasurfaces \cite{chen2023metawave,nolan2021ros}.

%% file: Chapter/system_design.tex
\section{System Design}

\begin{figure*}[htb!]
\includegraphics[viewport=-15 0 445 200, clip, width=\textwidth]{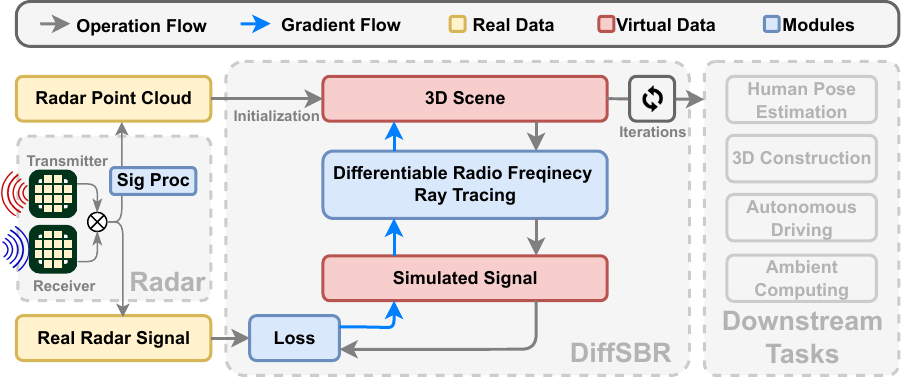} 
\vspace{-5pt} 
\caption{ Optimization begins with the raw radar signal; the signal is processed into point clouds for scene initialization. We then optimize the scene to generate a similar signal. During optimization, we use our differentiable radio frequency ray tracer, which allows both forward simulation and backpropagation of gradients.}
\label{fig:framework} 
\end{figure*}

\name adopts an iterative optimization framework for reconstructing 3D scenes from RF signals.  Figure \ref{fig:framework} illustrates its overall architecture and workflow. 
(i) It first initializes a parameterized 3D scene representation based on point clouds. 
(ii) The \textit{forward pass} involves differentiable RF ray tracing to simulate radar signals based on the generated 3D scene, incorporating RF material properties and multi-antenna Multiple Input Multiple Output (MIMO) arrays on the radar. 
(iii) To assess if the generated 3D scene matches the real scene as the output, the simulated signals are compared to observed/received signals using a spatial multi-antenna loss. 
(iv) Given this loss function, stochastic gradient descent is employed to guide the iterative optimization, update the 3D scene parameters, and minimize this loss. 
The gradients are computed by backpropagation through differentiable RF ray tracing. 
(v) After iterative optimization, the refined 3D scene parameters constitute the final reconstructed 3D representation that closely matches the true scene, as the sensing output, serving for the downstream tasks.

\subsection{3D Scene Representation and Initialization}
\label{sec:Initialization}
To reconstruct 3D scenes from a few observations using iterative optimization, we define 3D \textit{scene representation} and delineate the \textit{scene parameters} to be optimized. 
Our method encompasses a diverse set of 3D representations, including surface-based representations like triangle meshes, implicit models such as the signed distance field (SDF), and comprehensive volumetric approaches. It may also incorporate emerging representations with NeRF (Neural Radiance Fields)\cite{mildenhall2021nerf} and 3D Gaussians \cite{kerbl3Dgaussians}.
In general, any alternative 3D representations can be used in \name as long as they are compatible with ray tracing and differentiable with respect to their control parameters.

\vspace{-5mm}
\noindent \paragraph{Parameterization.}
Given the sparsity of mmWave signals, it is crucial to parameterize the scene for specific downstream applications, thereby reducing the optimization search space and lowering ambiguity. We consider the following mainstream mmWave sensing applications for case studies:

\textbf{\textit{(i) 3D Bounding Box Detection:}}  This is a primary application of mmWave sensing \cite{qian2021robust}. We adopt the transformation matrix of 3D mesh objects as the optimization parameter. Concurrently, positional encoding should be applied.

\textbf{\textit{(ii) Human Pose Estimation:}}  To estimate humans posture via mmWave radar, we employ the SMPL model \cite{loper2023smpl}, which incorporates 69 parameters to control human postures and another 10 parameters for body shape adjustments.

\textbf{\textit{(iii) Unseen Object Reconstruction:}} Voxel-based methods can be adopted for unseen object reconstruction, such as density field with triangulation. To mitigate ambiguity, one approach is to pre-train a voxel representation specifically for the target type of objects. Autoencoders can be employed to encode a high-dimensional voxel representation into a low-dimensional latent space, with subsequent optimizations performed within this latent space. The key advantages of this approach are the significant improvement in optimization efficiency and the reduction of ambiguity. Notably, training such a pre-trained encoder and decoder doesn't necessitate the collection of actual RF signal data, which can be labor-intensive and require specialized equipment. Instead, it suffices to leverage existing large-scale 3D model datasets.

\vspace{-5mm}
\noindent \paragraph{Initialization.}
Initialization from point clouds preprocessed from raw data has been previously demonstrated as an effective approach in prior works \cite{kerbl3Dgaussians}. 
\name can be initialized from these point clouds from mmWave radar. The initialization procedure can be based on registration techniques to compute the initial parameters for the aforementioned scene representation.

\subsection{Differentiable RF Ray Tracing}
\label{sec:Differentiable}

Given a 3D scene initialized and parameterized by a continuous set $\Theta$, which encapsulates elements such as radar pose, scene geometry, material properties, and dynamics, we need to generate the corresponding simulated radar signal $\mathcal{Y}$. 
Besides, considering a scalar function derived from this radar signal exists, such as a desired loss function to be optimized, another aim of this approach is to backpropagate the gradient of the scalar with respect to all scene parameters in $\Theta$.

Our Differentiable RF Ray Tracing is designed to achieve these dual tasks of forward simulation and backward propagation. 

\subsubsection{Ray Tracing}
\label{sec:rt}
\paragraph{Ray Tracing Forward Simulation.}
Ray tracing, an established technique in computer graphics, has been used in in computational RF to estimate parameters such as time of flight, velocity, and signal strength of electromagnetic radiation. 
Besides, sensing processing in RF can also be a similar function to rendering processing in graphics.
Therefore, inspired by the \textit{Rendering Equation} in graphics, we introduce an ``\textit{RF Rendering Equation}'' for RF sensing with Ray tracing to generate the simulated radar signal:

\begin{multline}
\small
S_r(d,\varphi_o)=\frac{P_t(\varphi_o)F(\varphi_o)G(\varphi_o)}{4\pi d^2} \text{PL}(d) \\ + \int_{\Omega} p(\mathbf{x})\bigg(\frac{P_t(\mathbf{x},\omega_i)F(\omega_i)}{4\pi|\mathbf{x}|^2} + S_r(\mathbf{x},\omega_i)\bigg)\cos\omega_i,d\omega_i,
\end{multline}

\noindent where \( S_r(d,\varphi_o) \) is the received signal power density at distance \( d \) and direction \( \varphi_o \). \( P_t(\cdot) \) signifies the transmitted power, with \( F(\varphi_o) \) and \( G(\varphi_o) \) denoting the directional gains of the transmitter and receiver, respectively. The term \( \text{PL}(d) \) represents path loss over distance \( d \). \( p(\mathbf{x}) \) is the reflection coefficient at position \( \mathbf{x} \), and \( \Omega \) encompasses the entire space of potential signal paths. \( \omega_i \) describes the solid angle of incident direction. The integral captures multipath contributions, with the recursive term \( S_r(\mathbf{x},\omega_i) \) representing multiple reflections akin to graphics' rendering equations.

To understand how the single propagates on each antenna within the RF rendering equation, ray tracing is used to simulate electromagnetic wave (i.e., mmWave signal) interactions with the generated 3D scene.  
Computing RF requires integration over all plausible RF paths perceived by antennas. This can be mathematically represented as:

\begin{equation}
\small
\bm{I} = \int_\mathcal{P} f(p, \boldsymbol{\Theta}) \, dp,
\end{equation}
\noindent where $f$ depends on scene parameters $\boldsymbol{\Theta}$ such as object positions, shapes, materials, etc., and $p$ denotes a ray path. However, solving this integral is often analytically and computationally intractable. Monte Carlo methods provide a statistical approach by taking random samples to approximate the integral:

\begin{equation}
\small
\bm{I} \approx \hat{I} = \frac{1}{N} \sum_{i=1}^{N} f(p_i, \boldsymbol{\Theta}),
\end{equation}
where $\hat{I}$ converges to $\mathbf{I}$ as $N \rightarrow \infty$. By leveraging Monte Carlo ray tracing, accurate RF channel characteristics can be efficiently simulated while avoiding expensive full-wave solutions of Maxwell's equations, especially in intricate environments with rich multipath reflections.

\paragraph{Ray Tracing Backpropagation.}
To achieve the backpropagation and calculate the partial derivative of each parameter of interest, Differentiating Ray Tracing is further designed, denoted as \( \bm{\theta} \in \bm{\Theta} \), with respect to the final output. The complexities arise due to the composition of both continuous and discontinuous integrands within function \( f \).

\textit{\textbf{Continuous Integrands:}} Most functions in RF ray tracing are continuous. These include the antenna radiation pattern, path losses, and reflection/transmission coefficients.  
One example is the attenuation function \( A(\cdot) \) that depends on the attenuation coefficient \( \theta_A \). Here, we decompose the function \( f \) into \( f'(\cdot) \) and \( A(\cdot) \), both of which are continuous with respect to \( \theta_A \):

\begin{equation} 
\small
    I \approx E = \frac{1}{N} \sum_{i=1}^{N} f'(p_i, \theta_A) \times A(p_i,\theta_A),
\end{equation}
Their partial derivatives can be calculated using automatic differentiation based on the chain rule:
\begin{multline}
\frac{\partial I}{\partial \theta_A} \approx \frac{\partial E}{\partial \theta_A} = \frac{1}{N} \sum_{i=1}^{N} \left( f'(p_i, \theta_A) \times \frac{\partial A(p_i,\theta_A)}{\partial \theta_A} \right. \\ \left. + A(p_i,\theta_A) \times \frac{\partial f'(p_i, \theta_A)}{\partial \theta_A} \right).
\end{multline}

\textit{\textbf{Discontinuous Integrands:}} 
Discontinuous integrals in ray tracing arise from visibility changes due to geometric edges and occlusion. To overcome this problem, we employ the reparameterization method \cite{loubet2019reparameterizing} that transforms non-differentiable integrals into differentiable ones using a change of variables.

Let $f(p,\theta)$ be a discontinuous integrand over $\mathcal{P}$, where $\theta$ denotes differentiable scene parameters. If a transformation $T: \mathcal{Q} \rightarrow \mathcal{P}$ exists, the integral can be reparameterized as:

\begin{equation}
\small
\int_{\mathcal{P}} f(p, \theta) dp = \int_{\mathcal{Q}} f(T(q, \theta), \theta) |\det J_T| dq,
\end{equation}

\begin{equation}
\frac{\partial I}{\partial \theta} = \int_{\mathcal{Q}} \left( \frac{\partial f}{\partial \theta} + f \frac{\partial}{\partial \theta}(\text{log}|\text{det}J_T|) \right) dq,
\end{equation}

\noindent where $J_T$ is the Jacobian of $T$. The key idea is to construct $T$ such that $f(T(q,\theta),\theta)$ no longer depends on $\theta$, enabling standard Monte Carlo integration and automatic differentiation. It is worth noting that while $T$ is designed individually for each integral with discontinuities, common transformations for vertices are already established in \cite{loubet2019reparameterizing,li2018differentiable}.

With this measure, the ray tracing components of the simulator become differentiable. We can then backpropagate the gradients from the ray tracing results, such as Time-of-Flights $I^t$ and signal strength $I^s$, to the input scene parameters $\mathbf{\Theta}$. The next step involves differentiating the RF component.

\subsubsection{RF Signal }
\label{sec:Simulated}
\paragraph{Simulated RF Signal Generation.}
After obtaining intermediate information $I$ from ray tracing, such as the time-of-flight $I^t$ and signal strength $I^s$, we can calculate the time-domain Intermediate Frequency (IF) signal to accurately simulate the mmWave signal. 
Note that the simulated IF signal follows the output format of a real radar, and can be represented as:
\begin{equation}
\small
S_{IF}(t) = \sum_{i=0}^{N} I^s_i \exp(2\pi j(\mu t I^t_i + f_c I^t_i)),
\end{equation}
where $N$ is the number of rays, $f_c$ is the carrier frequency, and $\mu$ is the frequency slope, given by $\mu = \frac{B}{T}$. $B$ denotes the signal bandwidth and $T$ represents the chirp duration. The terms $I^s_i$ and $I^t_i$ refer to the signal strength and time-of-flights of $i$-th path, respectively, derived from the ray tracing results.

\textit{\textbf{Material Properties:}}
To robustly simulate the signal and elevate the sensing performance, we model the electromagnetic material properties based on the Fresnel reflection coefficients derived from Maxwell's equations. The complex relative permittivity $\epsilon_r$ and permeability $\mu_0$ characterize each material.

The Fresnel reflection coefficients $r_p$ and $r_s$ for parallel and perpendicular polarizations depend on the incident angle $\delta_i$, transmission angle $\delta_t$, and wave impedance $\eta = \sqrt{\nicefrac{\mu_0}{\epsilon}}$ and complex permittivity $\epsilon = \epsilon_r \epsilon_0 - \frac{j \sigma}{\omega}$: 

\begin{equation}
\small
r_p = \frac{\eta \cos\delta_i - \cos\delta_t}{\eta \cos\delta_i + \cos\delta_t}, 
r_s = \frac{\cos\delta_i - \eta \cos\delta_t}{\cos\delta_i + \eta \cos\delta_t},
\end{equation}

\noindent where $\cos\delta_i$ and $\cos\delta_t$ are computed from the incident direction $\mathbf{i}$, surface normal $\mathbf{n}$, and relative permittivity $\epsilon_r$:

\begin{equation}
\small
\begin{aligned} 
\cos\delta_i &= -\textbf{i} \cdot \textbf{n},   & \sin\delta_i &= \sqrt{1 - \cos^2\delta_i}, \\ 
\sin\delta_t &= \sqrt{\epsilon_r} \sin\delta_i, & \cos\delta_t &= \sqrt{1 - \sin^2\delta_t}.
\end{aligned}
\end{equation}

\noindent This Fresnel model can help balance accuracy and efficiency for simulating complex RF propagation in our ray-tracing framework.  
The Fresnel reflection coefficients, which are differentiable with respect to the material properties (e.g., permittivity), are integrated into this framework. Every time a ray interacts with a surface, these coefficients are applied and accumulated, subsequently contributing to the signal strength $I^s$ of the path.

\paragraph{RF Signal Backpropagation.}
To backpropagate the gradient from the final signal to the ray tracing results, differentiation of the signal generation process is imperative. Fortunately, given that the signal generation is continuous, it is amenable to direct differentiation using automatic differentiation. Nevertheless, we also present our analytical differentiation approach:

\begin{equation}
\small
\frac{\partial S_{IF}(t)}{\partial I^t_i} = 2\pi j(\mu t + f_c) I^s_i \exp(2\pi j(\mu t I^t_i + f_c I^t_i)),
\end{equation}

\begin{equation}
\small
\frac{\partial S_{IF}(t)}{\partial I^s_i} = \exp(2\pi j(\mu t I^t_i + f_c I^t_i)).
\end{equation}

\subsection{End-to-End Backpropagation}

As described in Section \ref{sec:rt} and Section \ref{sec:Simulated}, we achieve both forward simulation and backpropagation of 3D scene $\Theta$ to path information $I$ and $I$ to simulated radar signal $S_{IF}(t)$. We can then achieve end-to-end backpropatation by calculating the partial derivative of simulated radar signal $S_{IF}(t)$ respects to all scene parameters $\Theta$ by using chain-rule or automatic differentiation:

\begin{equation}
\small
\frac{\partial S_{IF}(t)}{\partial \theta} = \sum_{i=0}^{N} \left( \frac{\partial S_{IF}(t)}{\partial I_i} \times \frac{\partial I_i}{\partial \theta} \right).
\end{equation} 

With the completion of the differentiable RF ray tracing, the simulator is now capable of not only accurately and efficiently simulating the radar signal based on input scene parameters but also backpropagating the gradient to all these parameters.

\subsection{Gradient-Based Optimization for 3D Scene}
\label{sec:Optimization}
\subsubsection{Optimization Formulation}
Given the radar-observed sparse signals $\bm{y}$ within a real-world scene, our goal is to generate a 3D digital reconstruction of the scene, denoted as $\bm{\theta}$. 
An ideal reconstruction would result in simulated radar signals from an RF simulator $S(\cdot): \Theta\to\mathcal{Y}$ that align closely with $\bm{y}$. 
Mathematically, the direct inversion, $\bm{\theta}\leftarrow S^{-1}(\bm{y})$, would provide the desired reconstruction. However, due to the complexity of $S(\cdot)$, obtaining a closed-form solution is not feasible.

To address this challenge, we introduce an iterative optimization framework that minimizes the discrepancy between the simulated signals $S(\bm{x})$ and the observed/received signals $\bm{y}$. This iterative process refines the scene parameters $\bm{x}_0$, leading to a more accurate 3D representation $\bm{x}^*$ of the real-world scene from the RF signals.  
The process can be expressed mathematically as: 

\begin{equation}
\small
\bm{\theta}^* = \arg\min_{\bm{\theta} \in \Theta} \ell(S(\bm{\theta}),y).
\end{equation}

To solve the optimization problem, we use Stochastic Gradient Descent (SGD). Specifically, the iterative update for our scene parameters using SGD is given by:
 
\begin{equation}
\small
\bm{\theta}_{t+1} = \bm{\theta}_{t} - \alpha_t \nabla \ell(S(\bm{\theta}_{t}),\bm{y}),
\end{equation}

\noindent where \( \alpha_t \) is the learning rate at iteration \( t \).
Using SGD, we iteratively refine the 3D scene representation by sampling subsets of the RF signals, computing the discrepancy gradients, and updating the parameters until convergence. Upon convergence, \name yields a 3D reconstruction \(\bm{\theta}^*\) that serves as a digital proxy for the real-world scene as the sensing output corresponding to the observed/received RF signals.

\subsubsection{Spatial Multi-Antenna Loss for Optimization}

The objective of radar-based 3D scene reconstruction is to identify the optimal scene parameters, $\theta^*$, which minimize the reconstruction loss, $\ell$, while adhering to constraints dictated by RF ray tracing. In contrast to standard camera imaging where each pixel corresponds to a single RGB value, radar systems involve antenna arrays that capture a temporal sequence of data. These sequences can consist of hundreds of thousands of values from RF signals, typically in the hundreds of megahertz range. This situation poses significant challenges since conventional loss functions may not be well-suited to the distinct properties of mmWave sensing data, such as high frequency and low sample count. For example, minor phase shifts might lead to substantial changes in Mean Squared Error (MSE) \cite{wang2009mean}, impeding the convergence of the model. Likewise, other loss functions like Kullback-Leibler (KL) \cite{kullback1951information} divergence face difficulties with high-frequency, low-sample data. Similarly, loss functions that rely on Fast Fourier Transform (FFT) images may not effectively capture fine details.

In the context of widely used multi-antenna MIMO radar, each transmitter-receiver antenna pair results in a unique signal, allowing the antenna array to intrinsically gather spatial information \cite{bekar2023enhanced,schroder2021experimental}. This aspect is crucial but often neglected in traditional temporal loss approaches applied to signal antennas, resulting in significant loss of information. To overcome this, we have developed a novel loss function that converts the native MIMO signals into a 3D spatial representation. We then employ an MSE-based criterion for optimization, leveraging the spatial information inherent in the MIMO radar data to its fullest extent.

\begin{equation}
\small
    \ell(y,\bar{y}) = \frac{1}{N} \sum_{i=1}^{N} (T(y)_i - T(\bar{y})_i)^2,
\end{equation}

\noindent where $T$ is a signal processing algorithm that maps raw MIMO signals to one 3D spatial image.

%% file: Chapter/implementation.tex
\section{Implementation}

\noindent
\textbf{mmWave radar platforms:}
We evaluate \name on 3 representative FMCW mmWave radar platforms: 
\textbf{(1) 2D ranging radar}: we employ an Infineon Position2Go module\cite{P2Go} as a ranging radar.
Position2Go operates on 24~GHz, with 1 TX and 2 RX antennas.
For each TX and RX pair, the raw in-phase and quadrature signals (I/Q) are accessible from a PC host connected to the radar.  
\textbf{(2) 3D automotive imaging radar}: we employ a TI AWR1843BOOST 76-81~GHz automotive radar \cite{AWR1843BOOST}, 
which has 3 TX and 4 RX antennas. 
It can output I/Q signals along with point cloud data, with 80 to 200 points per frame.  
\textbf{(3) 4D sensing radar}: 
we also test the 62-69~GHz Vayyar VtrigB \cite{vayyar_2022}, which 
features 4D radar sensing (distance, direction, relative velocity, and vertical).
VtrigB has 20 TX and 20 RX antennas, capable of producing I/Q signals and point clouds, with 1000 to 2000 points per frame, enabling simultaneous perception of multiple objects.

%% file: Chapter/experiment.tex
\section{Experimental Evaluation}

\begin{figure}
\vspace{-10pt}
\includegraphics[viewport=0 0 800 690, clip, width=0.46\textwidth]{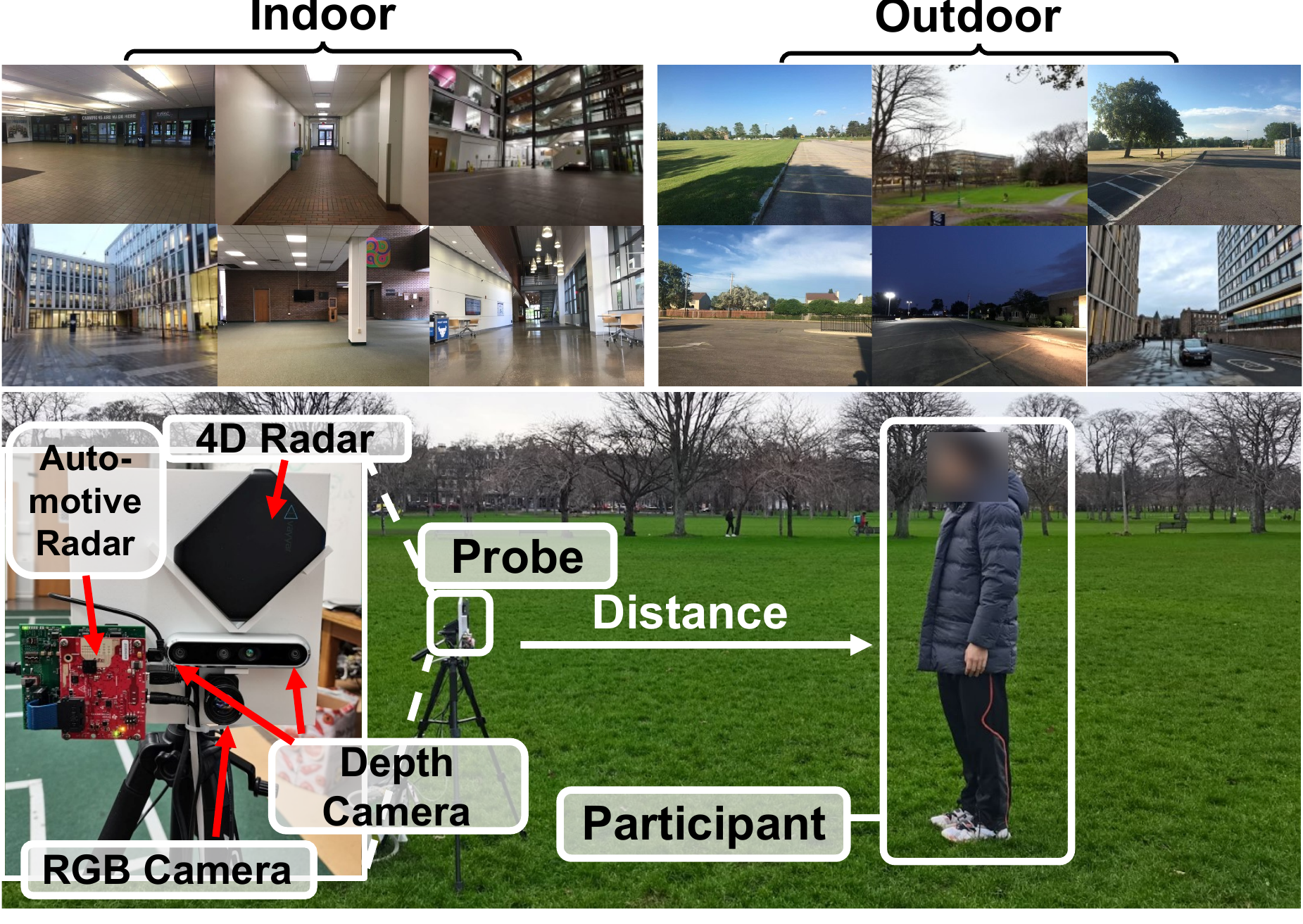}
\caption{System setup and test environments for \name.}
\vspace{-10pt}
\label{fig:envir}
\end{figure}

\begin{figure*}[htb!]
\includegraphics[viewport=-5 0 2500 1390, clip,width=\textwidth]{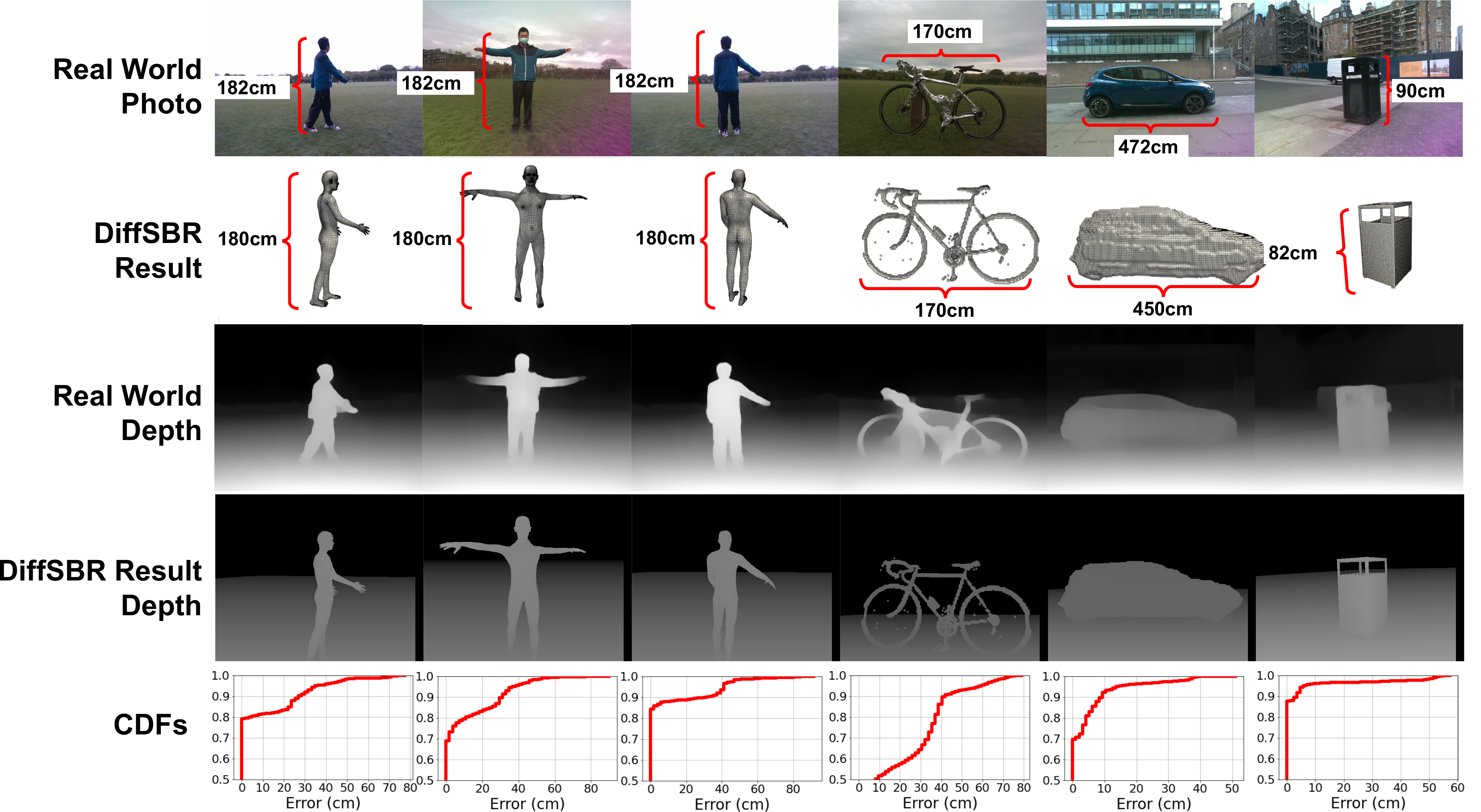}
\caption{
 Examples of  3D reconstruction results
}
\vspace{3mm}
\label{fig:radar_ti_result} 
\end{figure*}

\begin{figure*}[htb!]
\includegraphics[viewport =-120 0 2700 1756,clip, width=0.98\textwidth]{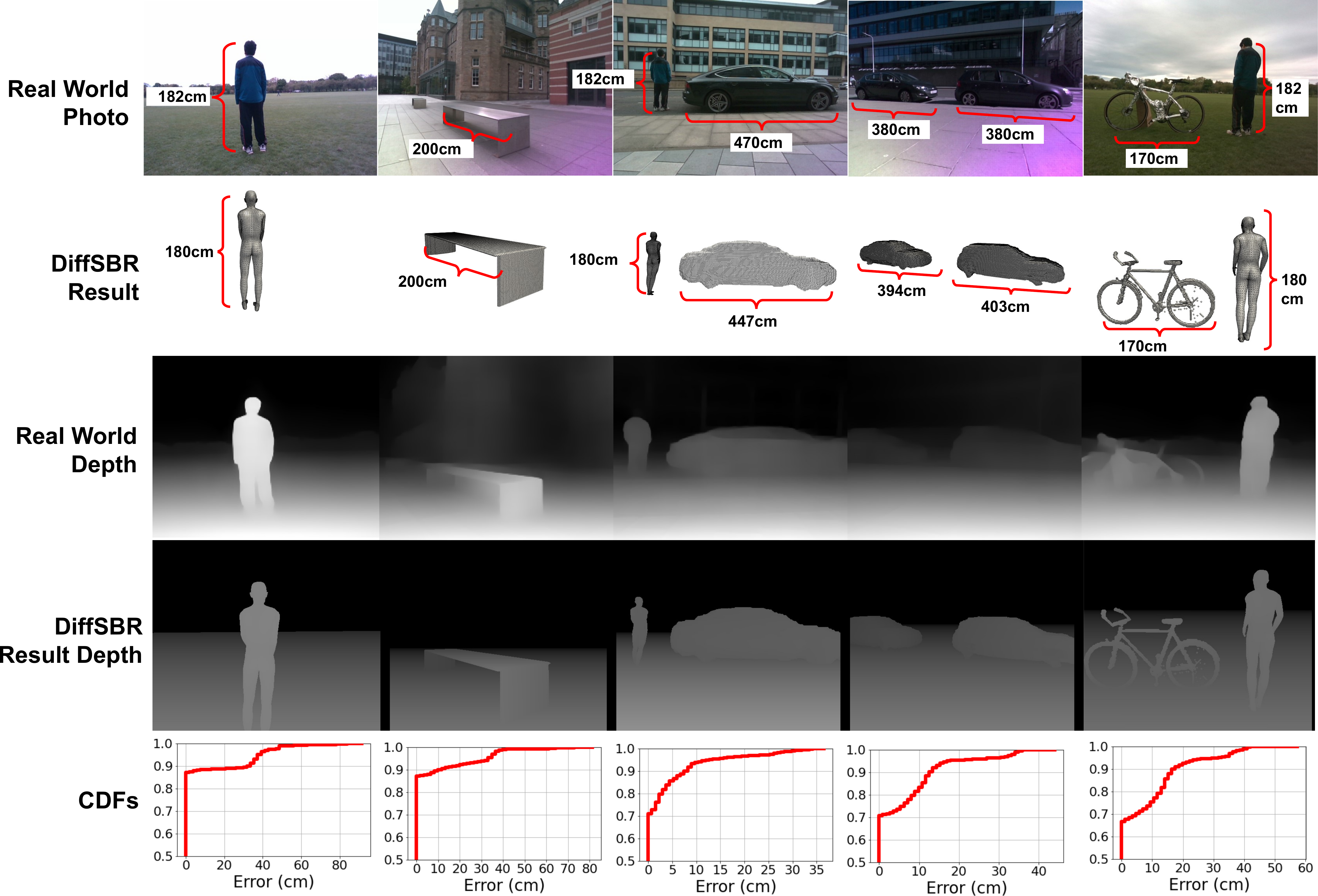}
\caption{Examples of  3D reconstruction results. 
}
\vspace{-6mm}
\label{fig:results_4d}
\end{figure*}

\begin{table*}[htb!]
\centering
\caption{3D Reconstruction Results on Multiple Objects.}
\begin{tabular}{lcccccc}
\toprule
            & Human & Bench & Car \& Human & Two Cars & Human \& Bike & Avg. \\ 
\midrule
Avg. SSIM  & 0.8884 & 0.9005 & 0.8655 & 0.9370 & 0.832 & 0.8798 \\
SD         & 0.0012 & 0.0037 & 0.0017 & 0.0023 & 0.0026 & 0.0023\\
HawkEye\cite{guan2020through} SSIM & 0.7231 & 0.6753 & 0.7456 & 0.6789 & 0.7765 & 0.7198\\
\bottomrule
\end{tabular}
\label{tab:ssim_radar_4D}
\vspace{-5mm}
\end{table*}

\subsection{Experimental Setup}
\label{sec:EvaluationSetup}
\noindent
\textbf{Ambient Environment:}
As shown in Figure~\ref{fig:envir}, we conduct experiments across 
6 indoor locations (\textit{e.g.}, classrooms, homes, halls) and 6 outdoor locations (\textit{e.g.}, outdoor campus, football fields, and parking lots). These locations represent a variety of ambient environmental structures and multipath conditions. 

\noindent
\textbf{Sensing Objects:}
Our experiments involve the following object categories corresponding to the target use cases in Section~\ref{sec:introduction}.
\textit{(1) Human} (for pedestrian sensing, vulnerable road user detection, posture reconstruction, etc.): 
We recruit 4 male and 3 female participants, with an average age of 25, and  
height ranging from 164cm to 183cm.  
\textit{(2) Cars} (for road object characterization, traffic/parking violation, etc.): We use 3 types of representative cars (hatchback, sedan, and SUV). 
\textit{(3) Multiple other objects} (for road hazard sensing, surveillance/perimeter security, etc.): We use an additional 5 object categories, including bikes, trash bins, and benches.

\noindent
\textbf{3D Scene Parameterization:}
\textit{(1) Human}: We adopt skeletal mesh to parameterize the human body, which can flexibly transform with multiple degrees of freedom around the joints. 
The body shapes are controlled by a binary (male/ female) along with 7 other parameters \cite{allen2003space}. 
Our actual test subjects can take any of the 14 most representative human poses, but the \name can accommodate arbitrary virtual poses.  
\textit{(2) Cars}: The shape of objects under the ``vehicle'' category varies greatly, e.g., hatchback, sedan, SUV, coupe, convertible, bus, and trucks. Therefore, we use a dynamic mesh to parameterize the vehicle 3D mesh. 
We define a 3D density field with a resolution of 512 sampling points per dimension, resulting in a total of 134 million parameters.
To ensure computational efficiency, we train a autoencoder on ShapeNet \cite{shapenet2015} to compress the density field to 16*16*16.
\name's optimizer adjusts these 4096 parameters, decompresses them back to 512*512*512, and then triangulates the density field to a mesh.  
\textit{(3) Bike and other objects:} As bicycles, benches, trash bins, and parking ticket machines do not undergo deformation themselves, the style and geometry are relatively centralized. Moreover, each object category usually follows a similar manufacturing standard \cite{bike_inc}.
Therefore, we use the static mesh directly to parameterize these objects.
The only parameter that determines the static mesh is the object type, which is used as an index to select candidates from the ShapeNet \cite{shapenet2015} model library.

\noindent
\textbf{Ground Truth:}
We use Intel Realsense D455 RGB and depth camera \cite{intel_realsense} to capture RGB and depth images as the ground truth for 3D reconstruction. 
The RGB sensor has a resolution of 640$\times$480, and ranges from 0.4~m to 10~m.  The depth sensing accuracy is around 5~mm.

\subsection{3D Reconstruction Performance}

\begin{table}[h!]
\vspace{-3mm}
\caption{3D Reconstruction Shape Results. }
\vspace{-5pt}
\small
\centering
\begin{tabular}{m{1cm}<{\centering}ccccc}
\toprule
     & Human & Bike & Car & Trash Bin & Avg. \\ 
\midrule

\textbf{Avg.SSIM} &  0.8492 &  0.8358 &  0.9183 &  0.9145  &   0.8772\\ \
\textbf{SD} &  0.0019 &  0.0007 &  0.0002 &  0.0298  &   0.0052\\ 
\bottomrule
\end{tabular}
\label{tab:ssim_radar_ti}
\vspace{-5pt}
\end{table}


\vspace{-5pt}

We evaluate the performance of \name in various scenarios. For the single-object 3D reconstruction, as shown in Table \ref{tab:ssim_radar_ti}, \name performs best on cars and trash bins, with an average SSIM of 0.92, demonstrating superior shape similarity with ground truth 3D meshes. Additionally, the size error rates are consistently below 5\% across different objects, with a depth accuracy of 85\% at a 20cm error tolerance, as demonstrated in Figure \ref{fig:radar_ti_result}.

When applied to multiple and complex object scenarios, as shown in Table \ref{tab:ssim_radar_4D}, \name achieves an average SSIM of 0.88. It again performs best on strong reflectors such as cars (average SSIM 0.93) and benches (average SSIM 0.9). 
In comparison with the {state-of-the-art} data-driven method HawkEye \cite{guan2020through}, our approach outperforms HawkEye across all object categories and significantly exceeds HawkEye's average SSIM of 0.72. 
This indicates not only superior accuracy but also improved generalization capabilities, achieved without the need for extensive pre-training on large datasets. 
Besdies, the depth accuracy from the \name 3D reconstruction is depicted in Figure~\ref{fig:results_4d} (bottom rows). 
For humans and car scenarios, the depth accuracy reaches above 85$\%$ when the error tolerance is 5~cm.
With two cars, the depth accuracy remains above 80$\%$ when the error tolerance is 10cm.
Overall, \name can reach a depth accuracy of 87$\%$ with an error tolerance of 20~cm across all the test scenarios. These results underscore the effectiveness of \name in achieving accurate 3D reconstructions across diverse object categories and scenarios, outperforming existing data-driven approaches and offering robustness in handling multiple and complex objects without needing extensive prior radar training data.

\subsection{Simulation Accuracy}
\begin{figure}[ht]
\includegraphics[viewport=13 0 660 330, clip,width=0.48\textwidth]{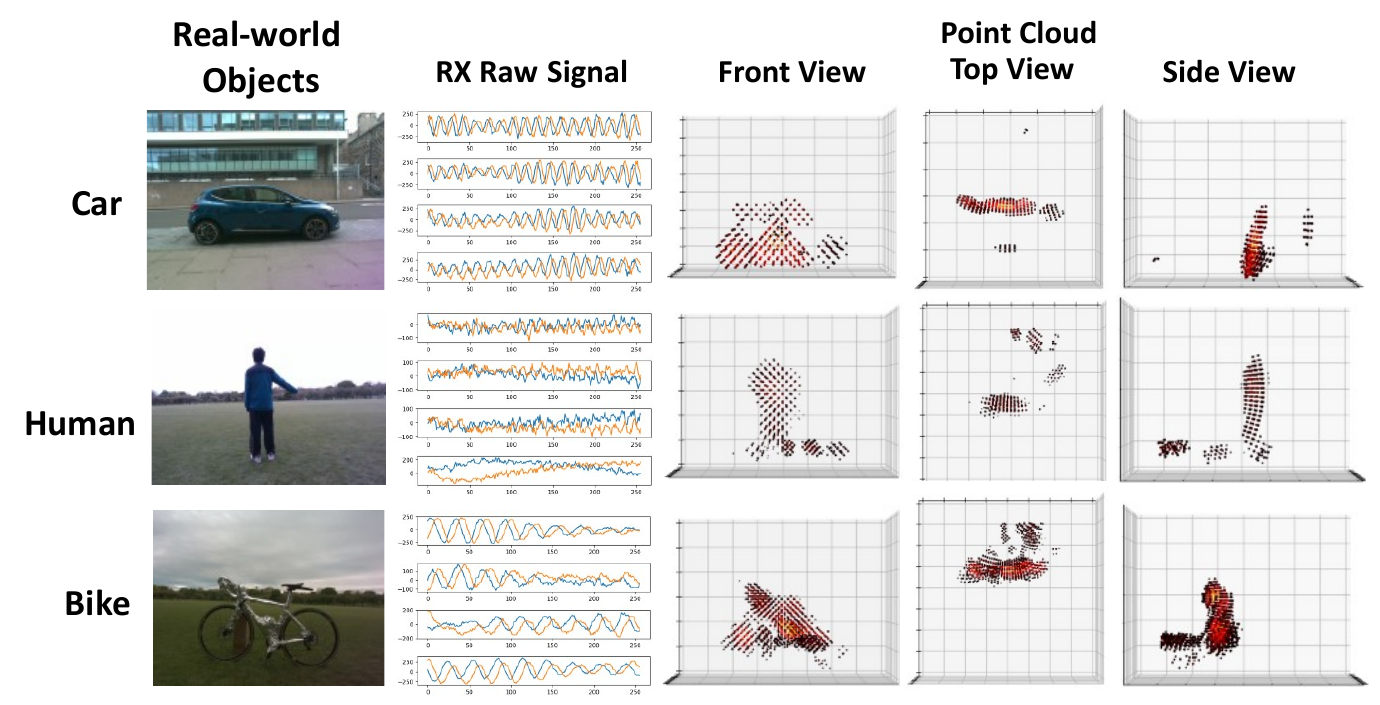}
\caption{Simulated received raw radar signal and the point cloud for example objects.  Signals from four RX antenna are illustrated, each with an I (blue) and Q (yellow) channel.} 
\label{fig:raw_signal}
\end{figure}
To verify whether \name can sense and reconstruct the objects following the physical laws correctly, we evaluate the RF simulator's capability and performance in the 3D scene compared with the ground truth. 
We used a total of 7 representative objects. Although \name uses a highly efficient yet simplified differentiable ray tracer, it consistently achieves an SSIM of around 0.99, in comparison to the electromagnetic field simulator. This proves that \name achieves high accuracy in its forward simulation process.

%% file: Chapter/discussion.tex
\section{Discussion}
We evaluate \name in representative multipath-rich practical environments, which aligns with almost all representative RF sensing work \cite{chen2023metawave,chen2020thermowave,xue2021mmmesh,guan2020through,ren20213d,qian20203d,jiang2020towards}.
The \name RF simulator thus only simulates one or more candidate objects, while omitting multipath reflections. 
Nonetheless, strong multipaths can cause a mismatch between the simulated and actual radar signals.
In the extreme case when the line-of-sight (LoS) is fully blocked (i.e., NLoS), the \name performance may degrade. 
It is still an open challenge for RF sensing under such a scenario.
To overcome this limitation, we can incorporate the ambient scenes into the 3D mesh, but this escalates \name's search space, making it intractably except in an environment with limited variability (e.g., road cross). 

%% file: Chapter/conclusion.tex
\section{Conclusion}
We proposed \name, a pioneering mmWave sensing paradigm fusing differentiable ray tracing with gradient-based optimization for robust 3D reconstruction. 
Central to \name is a unique differentiable RF simulator bridging the gap between sparse radar signals and detailed 3D representations. 
Experiments showcase \name's precision in characterizing object geometry and material properties, even for previously unseen radar targets. 
\name surpasses data-driven method limitations, exhibiting generalization across objects, environments, and radar hardware. 
\name revolutionized radio signal utilization and catalyzed advancements in computational sensing and computer vision cross-fields.